# A Coordination-based Approach for Focused Learning in Knowledge-Based Systems

## Abhishek Sharma

3840 Far West Blvd, Austin, TX 78731
Abhishek81@gmail.com

### Abstract

Recent progress in Learning by Reading and Machine Reading systems has significantly increased the capacity of knowledge-based systems to learn new facts. In this work, we discuss the problem of selecting a set of learning requests for these knowledge-based systems which would lead to maximum Q/A performance. To understand the dynamics of this problem, we simulate the properties of a learning strategy, which sends learning requests to an external knowledge source. We show that choosing an optimal set of facts for these learning systems is similar to a coordination game, and use reinforcement learning to solve this problem. Experiments show that such an approach can significantly improve Q/A performance.

## Introduction and Motivation

In recent years, there has been considerable interest in Learning by Reading [Barker et al 2007; Forbus et al 2007, Mulkar et al 2007] and Machine Reading [Etzioni et al 2005; Carlson et al 2010] systems. Rapid progress in these areas has significantly increased the capacity of learning systems to learn new facts from text. Optimal utilization of these facts could change the face of modern AI systems. However, we must make sure that the benefits of these new facts show up in better Q/A performance. Inundating a knowledge-base (KB) with irrelevant facts is hardly useful. Therefore, a rational learning system should try to formulate small number of learning requests which would help it to answer more questions. Which learning requests should be selected to maximize Q/A performance?

Techniques for selecting a small number of queries are also needed for active learning systems, which interact with a human expert or crowds to augment their knowledge. Since it would be impractical to seek thousands of facts from a single human user, learning systems must limit the scope of their queries. Even with crowdsourcing, selecting queries that avoid gathering irrelevant information is important.

In this work, we argue that an arbitrary selection of queries would result in large scale unification problems and the effect of new facts would not reach the target queries. We show that the selection of queries for maximizing Q/A performance is similar to a coordination game. We then use reinforcement learning to solve this problem. We model the dynamics of a learning system which sends learning requests to an external knowledge source, and experiments show that this coordination-based approach helps in improving Q/A performance. The results entail that the dependencies of search space induce a small partition (for each query) of the entire domain, which is selected by the reinforcement learning algorithm.

The rest of this paper is organized as follows. We start by discussing relevant work. After providing some background information, we discuss our learning model and its similarities to coordination games. We conclude after describing experimental results.

## Related Work

Our work is basically inspired by the idea that the deductive query graphs should be seen as a network of agents and coordination between them is needed for optimal results. The formal study of coordination in a network of agents has been done in social sciences [Jackson 2008], and theoretical computer science [Kleinberg & Raghavan 2005]. In [Kleinberg & Raghavan 2005], the authors have studied how game-theoretic ideas can be used for allocation of resources in query networks. In the AI literature, there has been some tangentially related work. For example, the relation between game theory and Boolean/first-order logic has been studied [Tang & Lin 2009, Dunne et al 2008]. Researchers in database community have discussed the importance of combining different constraints for optimal query plans [Hsu & Knoblock 2000]. QSAT problems have been seen as a two-person game [Kleinberg & Tardos 2005]. We have benefitted from these works because it is easy to see a deductive search space as a network. Although there has been work in the efficient deductive reasoning [Sharma et al 2016, Sharma & Goolsbey 2017, Sharma & Forbus 2010, Forbus et al 2009] , we are not aware of any prior work which has studied the similarity of coordination games and the dynamics of knowledge-based learning systems.

## Background

We use conventions from Cyc [Matuszek *et al* 2006] in this paper since that is the source of knowledge base contents used in our experiments[1]. Cyc represents concepts as *collections*. Each collection is a kind or type of thing whose instances share a certain property, attribute, or feature. For example, Cat is the collection of all and only cats. Collections are arranged hierarchically by the `genls` relation. (`genls <sub> <super>`) means that anything that is an instance of *<sub>* is also an instance of *<super>*. For example, (`genls Dog Mammal`) holds. Moreover, (`isa <thing> <collection>`) means that *<thing>* is an instance of collection *<collection>*. Predicates are also arranged in hierarchies. In Cyc terminology, (`genlPreds <s> <g>`) means that *<g>* is a generalization of *<s>*. For example, (`genlPreds touches near`) means that touching something implies being near to it.

### Coordination Games

Game theory has been extensively used for the study of interaction among independent, self-interested agents. The normal-form representation of a game is widely used to describe a game [Shoham & Leyton-Brown 2009]:

**Definition 1** (Normal-form game): A (finite, n-person) normal-form game is a tuple ($N$, $A$, $u$), where:

- $N$ is a finite set of n-players, indexed by $i$;
- $A = A_1 \times ... \times A_n$, where $A_i$ is a finite set of actions available to player $i$. Each vector $a = (a_1,...,a_n)$ is also called an action profile.
- $u = (u_1, ..., u_n)$, where $u_i : A \rightarrow R$ is a real-valued utility (or payoff) function for player $i$.

Coordination games are a restricted type of game in which the agents have no conflicting interests. In other words, it is the case that $u_i(a) = u_j(a)$ for any pair of agents $i$ and $j$. They can maximize benefits for all agents by coordinating their actions.

**Example:** A classic example of coordination game is the so-called Battle of Sexes. In this game, a husband and a wife wish to go to the movies, and they van select among two movies: "Lethal Weapon (LW)" and "Wondrous Love (WL)". They much prefer to go together rather than to separate movies, but while the wife prefers LW, the husband prefers WL. The payoff matrix is shown in Table

1 [Shoham and Leyton-Brown 2008]. We will return to this matrix later in the paper.

|       |    | Husband |     |
|-------|----|---------|-----|
|       |    | LW      | WL  |
| Wife  | LW | 2,1     | 0,0 |
|       | WL | 0,0     | 1,2 |

Table 1: The Battle of Sexes Payoff Matrix

### A Model for Learning Systems

Progress in deductive Q/A performance will require reasoning effectively with a continuously growing KB and $10^3$-$10^6$ first-order axioms. It is obvious that we would like to choose only those facts which are relevant for answering a given set of questions. Acquiring many irrelevant facts is inadvisable for two reasons: (a) Acquiring these facts from a human expert is expensive, and even crowdsourcing is not free and (b) These facts could affect the performance of deductive reasoning by increasing the number of failed reasoning chains. Therefore, large knowledge-based learning systems should prefer learning requests that ensure that they get only the most relevant facts from the external knowledge source. Very general queries like (`<predicate> ?x ?y`) would result in acquisition of very large number of facts. Therefore, in this work we use a learning model in which the learning requests are of the type (`<predicate> <C_1> <C_2>`), where $C_1$ and $C_2$ are specific collections. The set of specific collections consists of all collections which have less than N instances. In this work, N was set to 5,000. In what follows, we will discuss how the set of axioms used by the Q/A system plays an important role in determining the learning requests. The central task for the learning system is the following:

**Given**: N Learning Requests of the form (`<Predicate_i> C_11 C_12`)

**Task**: Find $C_{11}$, $C_{12}$, $C_{21}$, ..., $C_{N1}$, $C_{N2}$ such that they maximize Q/A performance.

A high-level view of our model of the learning system is shown in Figure 1. To simulate the learning behavior, we are using the inverse ablation model described in [Sharma & Forbus 2010]. The basic idea is to take the contents of a large knowledge-base (here, ResearchCyc) and make a simulation of the learning system by removing most of the facts. The operation of the learning component is simulated by gathering facts from the ablated portion of the KB that satisfy the learning requests, and adding them to the test KB. The initial KB consists of the basic ontology definitions (i.e., `BaseKB` and `UniversalVocabularyMt`) plus 5,180 facts selected at random. The rest of the ResearchCyc KB acts like an external knowledge source. At every time *t*, the learning system sends some queries to an external knowledge source. The answers are stored in KB(t) to get KB(t+1). For example, an example of such a learning request could be `(doneBy <Buying> <BritishCorporation>)`[2].

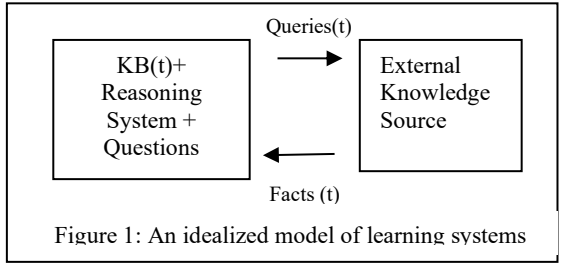

Figure 1: An idealized model of learning systems

Assume, for example, that we expect the learning system to find the country in which a person was born. In Figure 2, we show a simplified version of the search space which infers the country in which a person was born. The rectangular boxes represent AND nodes[3]. Note that the root node has been simplified in Figure 2; it actually refers to the formula `(holdsIn ?Birth-Event (objectFoundInLocation ?Person ?Location))`. For example, we may be expected to answer this query for the following set of people:

Q = {Einstein, Fermi, Riemann, Laplace, Hitler, Mao, Gauss, Feynman, Oppenheimer}

The variables in other nodes have not been shown for simplicity. To answer these queries and maximize the influence of axioms, the learning system needs facts involving the leaf nodes of the search space. Let us assume that the learning system decides to send following queries to the external knowledge source[4]:

```
t=0: (geographicalSubRegionsOfState <C1> <C2>).
t=1: (geoPoliticalSubDivisions <C3> <C4>)
t=2: (birthChild <C5> <C6>)
```

The aim of the learning system is to select the values of C1,…,C6 such that the Q/A performance is maximized. Let us consider two scenarios:

**Scenario 1:** `C1= US-State, C2= USCity, C3=AfricanCountry, C4=AfricanCity, C5= BirthEvent, C6= FrenchPhysicist.`

**Scenario 2:** `C1= US-State, C2= USCity, C3=US-State, C4=USCity, C5= BirthEvent, C6= USPhysicist.`

---

[2] This query represents all 'doneBy' statements involving buying actions done by British corporations.
[3] We are assuming an AND/OR query graph, where unification of antecedents is needed at AND nodes.
[4] We have ignored the `eventOccursAt` leaf node for simplicity.

Let KB(0) represent the contents of the initial KB. Let `Facts₁` and `Facts₂` represent the set of facts acquired from the external knowledge source in scenarios 1 and 2 respectively. Then KB(0) ∪ `Facts₁` would lead to little or no unification at the `spatiallySubsumes` and `objectFoundInLocation` queries. Very few answers would be inferred at the root node in this case because French scientists are generally not born in Africa. These problems would not arise in scenario 2 and many queries would be answered for the KB containing KB(0) ∪ `Facts₂`.

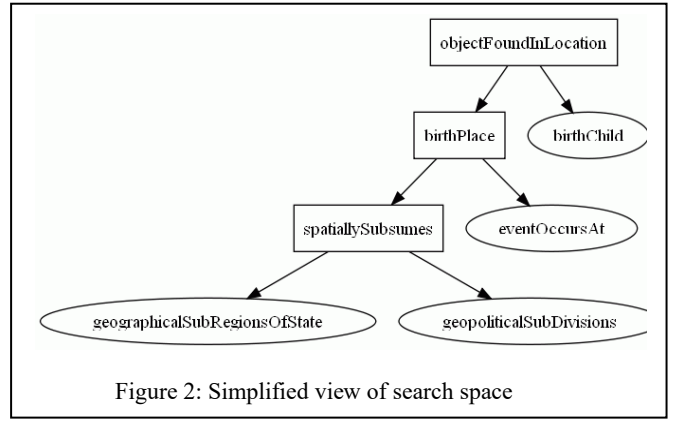

Figure 2: Simplified view of search space

The left child of the `objectFoundInLocation` node in Figure 2 infers the location of birth of the person, whereas its right child encodes the ethnicity of the person. As discussed above, lack of synchronization between what is inferred by different regions of the search space could lead to failures in inference. Let `?Location` and `?Ethnicity` be the variables which refer to what is inferred by the left and right child of the root node respectively. For simplicity, let us assume that the values of `?Location` and `?Ethnicity` are limited to French and American cities and scientists respectively. Table 2 summarizes the relation between Q/A performance and the choice of values for the variables. Similar problems would arise if there is more than one way for solving a query. We note that this explanation has been simplified in three ways: (a) The domains for these variables are much bigger, (b) Such tables can be made for every AND node in the search space, and (c) If we consider a bigger domain, we will see that the change in Q/A performance is not as abrupt as shown in Table 2, but we would observe a more gradual and graded change as the synchronization between different search branches improves.

|  | ?Location= < FrenchCity> | ?Location= <State-US> |
|---|---|---|
| ?Ethnicity = <FrenchPhysicist> | 42 | 0 |
| ?Ethnicity <USPhysicist> | 0 | 58 |

**Table 2: Relation between choice of learning queries and Q/A performance**

Notice the similarity between Table 1 and 2. In both cases, the top left and bottom right cells have significantly higher payoffs than other two cells. This variance arises due to the different expectations from different regions of the search space. In the example discussed above, we saw the possibility of spatial inconsistency between the outputs of different inference branches. Other types of inconsistencies (e.g., temporal inconsistencies) are also possible.

Now we can define the problem of coordination of learning actions in knowledge-based systems, to make the correspondences clear:

**Definition 2**: A normal-form game in a first-order learning system is a tuple ($N$, $A$, $u$), where:

- $N$ is a finite set of variables, indexed by $i$;
- $A = A_1 \times \ldots \times A_n$, where $A_i$ is a finite set of domains available to variable $i$.
- $u = (u_1, \ldots, u_n)$, where $u_i: A \to R$ is a real-valued utility (or payoff) function for player $i$.

We are assuming that the variable names are unique. In fact, there should be one agent/player for each argument position of predicate. For example, if (objectFoundInLocation ?person ?location) and (cityInCountry ?city ?country) are two nodes in the search space then:

N = {?person, ?location, ?city, ?country}
$A_1$ = {FrenchPerson, ChinesePerson, MuslimPerson, . }
$A_2$ = {EuropeanCountry, US-State, AfricanCountry, …}
$A_3$ = {USCity, AfricanCity, BritishCity , …}
$A_4$ = {AsianCountry, EuropeanCountry, …, …}

The $A_i$s are mapped to the learning requests. Their domains should be filtered to ensure that very general collections are excluded. The utility function maps the learning actions to the number of questions answered from facts acquired from their execution. Given this definition, we can say that the choice of optimal selection of learning queries in our model is similar to synchronization of actions in a coordination game.

This formulation implies that learning systems will need to find the optimal values of $C_i$s to maximize Q/A performance. When we choose values for these variables, we choose to reason about a small *context space* or a partition (e.g., Location of French Physicists, US Foreign Policy in Europe, etc.) of the entire domain.

## Reinforcement Learning

Reinforcement learning has been used for solving many problems, including game-theory. Our approach is based on the Joint Action Learner algorithm discussed in [Claus & Boutilier 1998, Bowling & Veloso 2002]. Their algorithm is aware of the existence of other agents and uses reinforcement learning to learn how different combination of actions affect the performance. Agents repeatedly play a

*stage game* in which they select an individual action and observe the choice of other agents. The net reward from the joint action is observed. This basically means that the environment of the repeated game is stationary (i.e., independent of time and history). However, the actions chosen by the agents are not independent of time and history. In fact, our aim is to show that *repeated games* would help the learning system to become cognizant of the interdependencies in the search space and guide the learning system to seek the optimal set of facts from the external knowledge source. The algorithm is shown in Figure 3.

The initial state of the system, $s_0$, is the initial state of the inverse ablation model (i.e., Ontology plus 5180 facts chosen at random). If the number of agents is $k$, and the size of each $A_i$ is $n$, then the total number of actions is m = $n^k$. Therefore, the set of all states is $\{s_0, s_1, s_2, \ldots, s_m\}$. For example, if $a_1 = (A_{11}, A_{21}, \ldots, A_{k1})$, then $s_1 = KB(0) + \cup_i Facts(A_{11})$. Here, Facts($A_{ij}$) represents the set of facts obtained from the external knowledge source when action $j$ is selected by agent $i$.[5]

---

Algorithm: Learning for player i.

1. Initialize Q arbitrarily, and for all $s \ \varepsilon \ S$, $a_{-i} \ \varepsilon \ A_{-i}$, set
   a. $C(s, a_{-i}) \leftarrow 0$, $n(s) \leftarrow 0$
2. Repeat,
   a. From state s select action $a_i$ that maximizes,
      $\sum_{a_{-i}} \frac{R}{n(s)}$,
      where, $R = C(s, a_{-i}) * Q(s, <a_i, a_{-i}>)$.

   b. Observing other agents' actions $a_{-i}$, reward $r$, and next state $s'$, set
      $Q(s, a) \leftarrow (1-\alpha) \ Q(s, a) + \alpha(r + \gamma V(s'))$
      $C(s, a_{-i}) \leftarrow C(s, a_{-i}) + 1$
      $n(s) \leftarrow n(s) + 1$

   where,
      $a = (a_i, a_{-i})$,    $V(s) = \max a_i \ \sum_{a_{-i}} \frac{R}{n(s)}$.

**Figure 3: Joint-Action Learning Algorithm** [Bowling &Veloso 2002, Claus & Boutilier 1998]

---

Recall that we have an agent for each argument position of every predicate, leading to a set of N agents and each agent $i$ can choose from a finite set of individual actions $A_i$. The chosen actions at any instance of the game are joint actions. The set of joint actions is $A_1 \times A_2 \times \ldots \times A_n$. The notation $A_{-i}$ refers to the set of joint actions of all agents excluding agent $i$ and $a_i$ refers to the action selected by agent $i$. In step 2b of the algorithm, the algorithm keeps a count of the number of times an agent has used a given action in the past. These relative frequencies provide

---

information about the agent's perception of the worth of different actions, and help the algorithm to have a model of the agent's current strategy. Each agent then uses them to choose a best response for the strategies of others [Claus & Boutilier 1998]. In step 2a of Figure 3, $C(s, a_{-i})/n(s)$ is the estimate that other agents would select the joint action $a_{-i}$ based on the history of play and their current strategy. Therefore, we choose action $a_i$ which would be the optimal response to this joint action.

## Experimental Analysis

Learning by Reading systems typically use a Q/A system to use what the system has learned. For example, Learning Reader used a parameterized question template scheme [Cohen *et al*, 1998] to ask questions. In this work, we have evaluated the efficacy of our approach on five types of questions used in Learning Reader. These question templates were: (1) Where did *<Event>* occur?, (2) Who was the actor of *<Event>*?, (3) Where might *<Person>* be?, (4) Who was affected by the *<Event>*?, (5) Where is *<GeographicalRegion>*? These questions were selected because the KB mainly contains information relevant for answering these questions. To test the performance for other predicates we also included a sixth query template `(temporallyIntersects ?x ?y)`. Since `temporallyIntersects` is a very general predicate, 2,038 other specializations are accessible through backchaining. In each template, the parameter (e.g., *<Person>*) indicates the kind of thing for which the question makes sense (specifically, a collection in the Cyc ontology). We use these questions in our experiments below, to provide realistic test of reasoning. When answering a parameterized question, each template expands into a set of formal queries, all of which are attempted in order to answer the original question. Each template contains one open variable, whose binding constitutes the answer. Our FIRE reasoning system uses backchaining over Horn clauses with an LTMS [Forbus & de Kleer 93]. We limit inference to Horn clauses for tractability. We use network-based optimization techniques for automatically selecting an efficient set of axioms. Inference is limited to depth 5 for all queries, with a timeout of 90 seconds per query. As shown in Fig 1, we are using the inverse ablation model. We start with KB(0) and use two strategies (i.e., baseline and coordination-based) for getting new facts from the external knowledge source. In this paper we have argued that learning systems would be able to improve their Q/A performance by using coordination algorithms. Therefore, for the baseline, we are using an approach which is oblivious of the presence of other agents/nodes. In other words, the baseline algorithm chooses the learning goal which had yielded the highest number of facts in the past (plus some exploration). We compare the performance of this baseline with the coordination algorithm shown in

Figure 3 which is cognizant of the presence of other nodes and tries to learn the effect of dependencies in the search space. To ensure adequate exploration, the agents chose a random action with probability 0.05. The learning rate, $\alpha$, was set to 0.5. We report results for the six question types mentioned above. The results are shown in Table 3.

The method proposed performs better for Experiments 1, 5 and 6 because their answers are derived from denser regions of KB. (In this context, density is calculated by averaging the number of facts per entity.) In such cases, a coordination-based approach is more likely to be useful due to higher probability of unification. It is clear that it will be difficult to find the optimal solution if the domains, i.e. $A_i$s, are large. Complete discussion of this issue would require another paper. Our preliminary studies show that the best performance would be obtained from collections of medium size.

| Query Type | Algorithm | No. of Queries | No. of Ans. | Improvement w.r.t. Baseline |
|---|---|---|---|---|
| 1 | Baseline | 13,153 | 4,878 | - |
| | Coordination | 13,153 | 8,487 | 74% |
| 2 | Baseline | 13,153 | 1,104 | - |
| | Coordination | 13,153 | 1,556 | 41% |
| 3 | Baseline | 5,299 | 249 | - |
| | Coordination | 5,299 | 315 | 26% |
| 4 | Baseline | 13,153 | 510 | - |
| | Coordination | 13,153 | 627 | 23% |
| 5 | Baseline | 33,915 | 4,856 | - |
| | Coordination | 33,915 | 12,233 | 151% |
| 6 | Baseline | 36,564 | 9,211 | - |
| | Coordination | 36,564 | 15,658 | 70% |

**Table 3: Experimental Results**

## Conclusions and Discussion

Large knowledge-based systems often need to identify a set of learning requests which could be answered with the help of an external knowledge source. Dead-end reasoning paths and expensive user time are two reasons for investigating this problem. We have shown that this problem is similar to a coordination game and reinforcement learning can be used for solving this problem. This approach finds a small partition which is induced by different expectations from different branches of the search space. Queries from this partition become learning requests. Experiments show that this coordination-based approach helps in improving Q/A performance.

Can the problem of synchronization be avoided if our learning requests are not of the type `(predicate <Collection> <Collection>)`? We do not believe so. Recall that the problem arises due to two reasons: (a) From our inability to get all relevant facts from a human expert, and (b) Increase in the number of failed reasoning chains due to irrelevant facts. These problems would continue to arise even if we use an entity-based learning strategy[6]. For example, a Machine Reading or learning by reading system can easily gather thousands of facts about topics like `UnitedStatesOfAmerica` from the Web. We cannot guarantee that all these facts would be useful because the utility of facts is determined by the set of axioms and their expectations. This implies that getting a smaller set of facts about more specific sub-topics (e.g., "Foreign Policy of United States") could lead to better and more efficient Q/A performance if they combine well with other expected queries. Similarly, it would be better to design specific and synchronized queries about entities for a human expert than to acquire small number of disconnected facts about a bigger topic.

Can we avoid this problem by using a centralized topic selection algorithm? Note that there is no simple mapping between the topic to the optimal set of learning queries. For example, if the learning system chooses to learn about the US economy, there is no obvious method which could tell us which aspect of this complex problem would be most useful for answering a given set of questions[7]. In fact, the optimal selection of actions in the coordination game discussed above maps the topic to the best set of learning requests. These results suggest three lines of future work. First, due to semi-decidable nature of first-order reasoning with Horn axioms, it is difficult to determine if we have achieved the best possible performance. Therefore, we would like to see if an algorithm like WoLF[Bowling & Veloso 2002] can further improve the performance. Secondly, we have noticed huge variance in the final solution even when values of most variables are fixed. This shows that some variables have more pivotal position and wield more influence in the search space. We would like to find if structures like "backbones" and "backdoors" (as discussed in SAT literature) can be identified here. Finally, we plan to implement these techniques in a real learning system and study its performance. The aim of this paper is to identify similarities between game theory and deduction in first–order reasoning systems. It is hoped that this work will stimulate further research in identifying other similarities between these domains.

---

Symposium on Learning by Reading, 2009